\pdfoutput=1

\documentclass[11pt]{article}

\usepackage[]{emnlp2021}

\usepackage{amssymb}   %
\usepackage{amsmath}   %
\usepackage{amsfonts}  %
\usepackage{amsthm}    %
\usepackage{latexsym}  %
\usepackage{stmaryrd}  %

\usepackage{graphicx}  %
\usepackage{subcaption}
\usepackage{color}     %
\usepackage{xcolor}    %

\usepackage{verbatim}  %
\usepackage{listings}  %
\usepackage{enumitem}  %
\usepackage{algorithm} %
\usepackage[noend]{algpseudocode} %

\usepackage{url}       %
\usepackage{xspace}    %
\usepackage{etoolbox}  %
\usepackage{hyperref}

\usepackage{array}     %
\usepackage{multirow}  %
\usepackage{multicol}  %
\usepackage{hhline}    %
\usepackage{dcolumn}   %
\usepackage{booktabs}  %

\usepackage{tikz}      %
\usepackage{pgfplots}  %
\pgfplotsset{compat=1.14}

\usepackage{times}     %
\usepackage[T1]{fontenc}       %
\usepackage[utf8]{inputenc}    %
\usepackage{microtype} %

\newcommand{\sepout}{\texttt{@\!@}}
\newcommand{\sepin}{\texttt{\#\!\#}}
\newcommand{\casper}{CASPER\xspace}
\newcommand{\casperOrig}{CASPER\textsubscript{orig}\xspace}
\newcommand{\casperOrigAbbr}{C\textsubscript{o}\xspace}
\newcommand{\casperAnon}{CASPER\textsubscript{anon}\xspace}
\newcommand{\casperfull}{ControllAble Semantic Parser via Exemplar Retrieval\xspace}

\newcommand{\Dtrain}{\ensuremath{\mathcal{N}_\text{train}}}
\newcommand{\Dsupport}{\ensuremath{\mathcal{N}_\text{sup}}}
\newcommand{\Ddev}{\ensuremath{\mathcal{N}_\text{dev}}}
\newcommand{\Xtrain}{\ensuremath{\mathcal{O}_\text{train}}}
\newcommand{\Xdev}{\ensuremath{\mathcal{O}_\text{dev}}}

\title{Controllable Semantic Parsing via Retrieval Augmentation}

\author{Panupong Pasupat \and Yuan Zhang \and Kelvin Guu \\
Google Research \\
{\tt \{ppasupat,zhangyua,kguu\}@google.com} \\}

\begin{document}
\maketitle
\begin{abstract}
In practical applications of semantic parsing,
we often want to rapidly change the behavior of the parser, such as enabling it to handle queries in a new domain, or changing its predictions on certain targeted queries.
While we can introduce new training examples exhibiting the target behavior,
a mechanism for enacting such behavior changes without expensive model re-training would be preferable.
To this end, we propose \casperfull (\casper).
Given an input query, the parser retrieves related exemplars from a retrieval index, augments them to the query, and then applies a generative seq2seq model to produce an output parse.
The exemplars act as a control mechanism over the generic generative model: by manipulating the retrieval index or how the augmented query is constructed, we can manipulate the behavior of the parser.
On the MTOP dataset, in addition to achieving state-of-the-art on the standard setup,
we show that \casper can parse queries in a new domain, adapt the prediction toward the specified patterns, or adapt to new semantic schemas without having to further re-train the model.

\end{abstract}

\section{Introduction}\label{sec:introduction}

\begin{figure}
\centering
\includegraphics[width=\linewidth]{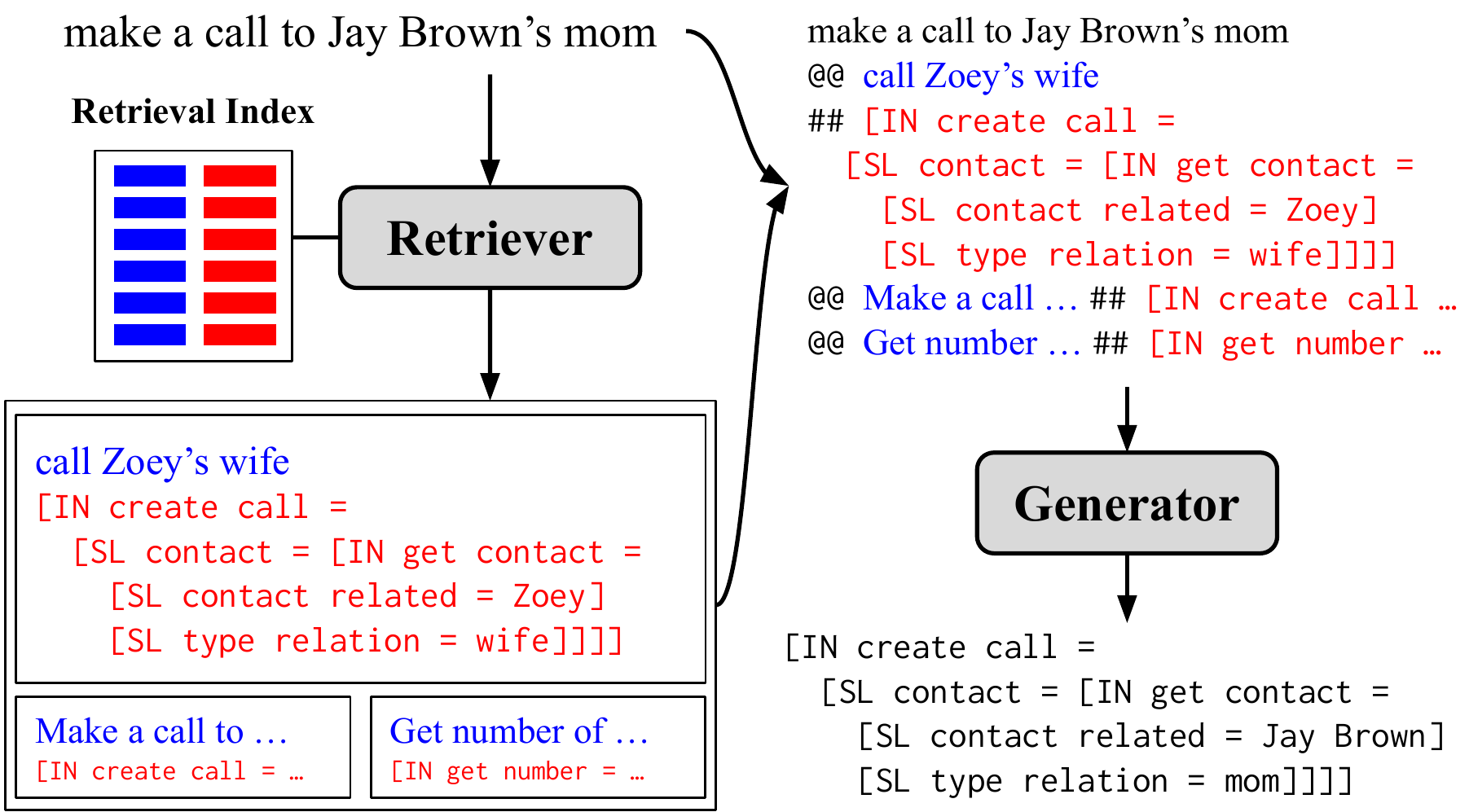}
\caption{The \casper model. (1) Given a query $x$, we retrieve exemplars $(x'_i, y'_i)$ from the retrieval index. (2) We construct an augmented query $x^+$ based on $x$ and the retrieved exemplars. (3) We apply a generative model on $x^+$ to produce an output parse $y$. The retrieval index and augmentation procedure can be modified to change the parser's behavior without re-training.
}
\label{fig:pull-figure}
\end{figure}

Semantic parsing is the task of mapping input queries to their meaning representations.
In practical applications of semantic parsing,
such as conversational agents,
we often want to rapidly \emph{control} the behavior of the parser.
We particularly focus on three scenarios:
(1) \textbf{Domain bootstrapping:} making the parser handle queries in a new domain \cite{su-yan-2017-cross,hou-etal-2020-shot,li-etal-2021-shot}.
This requires predicting new semantic labels (e.g., intents and slots) unseen during training, and assigning correct values to such labels.
(2) \textbf{Parse guiding:} influencing the prediction toward a specific parse pattern.
This can be used as an override for sensitive queries or queries that the model struggles on.
(3) \textbf{Schema refactoring:} adapting the parser to changes in the semantic schema such as semantic label renaming \cite{Gaddy2020OvercomingCD}.

A common way to control the parser's behavior is to construct training examples exhibiting the new behavior (e.g., examples from the new domain) and tune the model on them. %
However, model training requires computational resources, which can become unwieldy if we need to make multiple rapid changes.
Ideally, we want to control the behavior of the semantic parser \emph{without additional training}. Such an ability would enable many novel use cases.
For example, developers could update the semantic parser's behavior and observe the results immediately, thus speeding up the development cycle.
This can be used to quickly update the parser in time-critical scenarios while waiting for a fully re-trained model.
Another use case is deploying a single model to service multiple clients. Each client can manipulate the parser to fit its use without interfering with the central model or other clients, thus saving resources and preserving privacy.

To this end,
we propose \casperfull (\casper).
As illustrated in Figure~\ref{fig:pull-figure}, the parser first retrieves \emph{exemplars} relevant to the input query (e.g., training examples resembling the input query) from a modifiable retrieval index.
These retrieved exemplars are then augmented to the query. Finally, a seq2seq generator model takes the augmented query as input and generates a meaning representation.
The model takes inspiration from recent works that use modifiable retrieval indices \cite{Khandelwal2020GeneralizationTM,Khandelwal2021NearestNM} and exemplar-augmented inputs \cite{Brown2020GPT3,Liu2021WhatMG}.

The retrieval and augmentation processes grant us several ways to control the behavior of the parser. For instance, in domain bootstrapping, we can add examples from the new domain to the retrieval index. When these added examples are retrieved, the generator can condition on them while generating the output. This allows the generator to, for instance, follow the semantic template of the exemplars and produce new semantic labels unseen during training.

We evaluate our approach on the English portion of the MTOP dataset
\cite{li-etal-2021-mtop}.
In our experiments, we show that \casper preserves the generality and increases the performance of a seq2seq-based semantic parser, while also enabling new capabilities that are simply not possible with standard seq2seq parsers.
Our main results are:\footnote{The code for the experiments is available at \url{https://casperparser.page.link/code}}

\begin{itemize}
\item \textbf{Standard setup:}
On the English portion of MTOP, \casper
gives 2.1\% absolute improvement over the existing state-of-the-art, and 1.3\% improvement over the underlying T5 seq2seq parsing model.
\item \textbf{Domain bootstrapping:} By adding examples from a new domain to the retrieval index at test time, \casper can parse examples in the new domain without model re-training, while also preserving performance on other domains.
\item \textbf{Parse guiding:} We can train \casper to follow the semantic template of the manually provided exemplars when asked to do so, while maintaining accuracy on the standard setup.
\item \textbf{Schema refactoring:} By editing the retrieval index, \casper can, without re-training, adapt to a new semantic schema where some semantic labels are split into unseen labels.
\end{itemize}

\section{Approach}\label{sec:approach}

We present \casperfull (\casper) for parsing queries $x$ into meaning representations (MRs) $y$. As demonstrated in Figure~\ref{fig:pull-figure}, the prediction procedure consists of the following steps:

\begin{enumerate}
\item \textbf{Exemplar retrieval:} Retrieve a list $E$ of $k$ \emph{exemplars} $(x'_i, y'_i)$ ($x'_i$ is a query; $y'_i$ is the MR of $x'_i$) that are related to the input query $x$.
\item \textbf{Augmentation:} From $x$ and $E$, construct a retrieval-augmented query $x^+$.
\item \textbf{Generation:} Use a generative seq2seq model to map $x^+$ into an output MR $y$.
\end{enumerate}

We will elaborate on each step in the following subsections.
We can view exemplar augmentation as a way to provide extra information to any seq2seq-based semantic parser, while still preserving its ability to generate complex outputs.
The generator can learn to consider or ignore the provided exemplars, so \casper can perform at least as well as the underlying generator in the standard setup (Section~\ref{sec:standard-setup}).
Additionally, we will later show that by manipulating the retrieval index and how the augmented queries are constructed, we can control the behavior of \casper for
domain bootstrapping (Sections~\ref{sec:domain-bootstrapping}), parse guiding (Section~\ref{sec:parse-guiding}),
and schema refactoring (Section~\ref{sec:schema-refactoring}).

\paragraph{Retrieval}
The retrieval index consists of input-output pairs $(x', y')$, and is initially constructed from training examples.
We utilize a retriever that uses query embedding cosine similarity as the retrieval score.
Concretely, each index entry $(x', y')$ is indexed with the embedding $e(x')$ of the query, computed using a sentence embedder. Given an input $x$, we rank all index entries $(x', y')$ by the cosine similarity between $e(x)$ and $e(x')$, and let the list $E$ of exemplars be the top-$k$ entries.

For our experiments, we use the large version of the pre-trained Universal Sentence Encoder (USE-large) \cite{Cer2018UniversalSE} to embed the queries. We did not fine-tune the embedder.
As the retrieval index is small enough ($\approx$16k entries), we simply rank all index entries and choose $k = 5$ top entries as the exemplars.

\paragraph{Augmentation}

From the input query $x$ and the list $E = [(x'_1, y'_1), \dots, (x'_k, y'_k)]$ of retrieved exemplars, we construct an augmented query $x^+$.
Similar to previous works that use exemplar-augmented inputs \cite{guu-etal-2018-generating,Lewis2020RAG,Brown2020GPT3,Liu2021WhatMG},
we simply concatenate each exemplar to the query:
\[
x^+ = x \;\sepout\; x'_1 \;\sepin\; y'_1 \;\sepout\; x'_2 \;\sepin\; y'_2 \;\sepout\; \dots
\]
where \sepout{} and \sepin{} are the separator strings. The MRs $y'_i$ are simply treated as strings.

\paragraph{Generation}
We fine-tune a pre-trained seq2seq model to map the augmented query $x^+$ to the string representation of $y$. For our experiments, we fine-tune T5-base \cite{Raffel2020T5}.
While T5 was pre-trained on text data, our experiments show that it can effectively generate MRs after fine-tuning.

\paragraph{Training}
We keep the retriever fixed and only train the generator model.
When constructing $(x^+, y)$ pairs for training the generator, we want to \emph{diversify} the list of exemplars $E$. This would encourage the generator to learn when to use or ignore the exemplars based on the their quality and relevance to the input $x$.
To this end, instead of using top-$k$ retrieval results as exemplars, at training time we construct a \textbf{sampled-$k$} exemplar list $E$ as follows. From the input $x$, we first 
create a ranked pool of all index entries, excluding ones whose query is exactly $x$.
At each step $i \in \{1,\dots,k\}$, we choose the $j$th entry from the pool with probability $\propto p(1-p)^{j-1}$ (where $p$ is a hyperparameter, set to 0.5 in the experiments). This geometric distribution makes higher-ranking entries get sampled more frequently. We then remove the sampled exemplar from the pool and add it to $E$. %

\subsection{Faithfulness toward exemplars}\label{sec:faithfulness}

Although the generation of $y$ is conditioned on the exemplars in $E$, the generator could implicitly ignore the exemplars entirely. This is desirable as it allows the model to generate reasonable outputs even when the exemplars are of low quality.
However, if the parser always ignores the exemplars, we will not be able to control the parser via exemplar manipulation, and the parser might struggle on out-of-distribution examples at test time (e.g., in the domain bootstrapping setup).

We want the parser to be more faithful toward exemplars, but still want the generator to make a judgment call to ignore the exemplars when appropriate. Additionally, we want an  on-demand mechanism for adjusting the degree of faithfulness on specific queries. We thus propose the following techniques:

\begin{figure}
\centering{\small
\begin{tabular}{@{}r@{ }l@{}}
& \textbf{Original exemplars and target output:} \\
$y'_1$: &
[IN create call = [SL contact = [IN get contact = \\
& \quad [SL contact related = Zoey]  [SL relation = wife]]]] \\
$y'_2$: & [IN get call = [SL contact = Jack] [SL time = today]] \\
$y$: &
[IN create call = [SL contact = [IN get contact = \\
& \quad [SL relation = dad]]] [SL name app = Whatsapp]] \\
& \textbf{Anonymized:} \\
$y'_1$: &
[IN 42 = [SL 39 = [IN 53 = \\
& \quad [SL 6 = Zoey]  [SL 94 = wife]]]] \\
$y'_2$: &
[IN 12 = [SL 39 = Jack] [SL 71 = today]] \\
$y$: &
[IN 42 = [SL 39 = [IN 53 = \\
& \quad [SL 94 = dad]]] [SL 88 = Whatsapp]] \\
\end{tabular}
}
\caption{Anonymized exemplars and target output.}
\label{fig:improving-faithfulness}
\end{figure}

\paragraph{Anonymization}
Most seq2seq models, including T5, can learn to copy\footnote{T5 does not have an explicit copy mechanism, but it effectively learns to produce the same tokens as the input.} tokens from the input string. However, with regular supervision, the model may still end up not learning to copy semantic labels from the exemplars if
(1) such labels appear so frequently that the model memorizes their usage and generates them without copying, or (2) the retrieval is imperfect and copying from exemplars hurts the model during training.

To explicitly teach the generator to copy labels from the exemplars,
we create additional \emph{anonymized training data} where each unique semantic label in $y'_i$ and $y$ are turned
into a random numerical label, as illustrated in Figure~\ref{fig:improving-faithfulness}.
Since the labels are anonymized differently in each example, the generator
can no longer memorize their usage, and
must learn to identify and copy the correct anonymized labels.
We train the generator on an equal mix of original and anonymized data.

\paragraph{Guiding tag}
In some scenarios,
we want to manually instruct the model to be more faithful toward the exemplars than usual. To do so, we utilize a special token (``PLATINUM'' in our experiments) as a \emph{guiding tag}. We simply insert $T$ before each exemplar when constructing $x^+$: 
$$
x^+ = x \;\sepout{}\;
T\; \tilde x_1 \;\sepin{}\; \tilde y_1\;\sepout{}\;
T\; \tilde x_2 \;\sepin{}\; \tilde y_2\;\sepout{}
\dots
$$

To establish the behavior of the guiding tag,
we create additional training examples $(x^+, y)$ where $x^+$ contains the guiding tag, and the prediction $y$ is considered highly faithful to the augmented exemplars in $x^+$. One instantiation, \emph{oracle training data}, can be constructed by constraining the retrieved exemplars $(x'_i, y'_i)$ so that $y'_i$ and $y$ share the \emph{semantic template} (any notion of semantic similarity; e.g., the MR's labels and hierarchical structure).
The generator is trained on the combination of this oracle training data and the normal data.
\section{Standard setup experiments}\label{sec:standard-setup}

\begin{table}[t]
\centering
{\small
\begin{tabular}{@{}lc@{\;\;}cc@{\;\;}c@{}}
\toprule
& \multicolumn{2}{@{}c@{}}{\textbf{Dev}}
& \multicolumn{2}{@{}c@{}}{\textbf{Test}} \\
\textbf{Method} &
\textbf{Exact} &
\textbf{Template} &
\textbf{Exact} &
\textbf{Template} \\
\midrule
mBART+MT %
& - & - & 84.3\phantom{0} & -  \\
T5 
& 83.18 & 87.22
& 85.06 & 88.70 \\
\casper
& 84.29 & 87.65
& 85.54	& 89.13 \\
\casperOrig
& \textbf{84.67} & \textbf{87.98}
& \textbf{86.36} & \textbf{89.65} \\
\casperAnon
& \underline{79.61} & \underline{82.60}
& \underline{80.85} & \underline{83.90} \\
\bottomrule
\end{tabular}
}
\caption{\textbf{Standard setup:} exact match and template accuracy on the English portion of MTOP. \casper outperforms T5 and previous state-of-the-art. (\underline{underlined} = worse than baseline)}
\label{tab:standard-results}
\end{table}
\begin{figure}[t]
\centering
{\small
\begin{tabular}{@{}r@{ }l@{}}
\toprule
$x$: & What's the biggest story today? \hfill \textbf{(a)} \\
$x'_1$: & what's the top story for today? \\
$y'_1$: & [IN get stories news = \textcolor{green!50!black}{[SL news reference = top]} \\
& \quad [SL news type = story] [SL date time = for today]] \\
$x'_4$: & Tell me the biggest news story of the day. \\
$y'_4$: & [IN get stories news = [SL news type = news story]] \\
\midrule
T5: & [IN get stories news = [SL news type = story] \\
& \quad [SL date time = today]] \\
\casperOrigAbbr: & [IN get stories news = \textcolor{green!50!black}{[SL news reference = biggest]} \\
& \quad [SL news type = story] [SL date time = today]] \checkmark \\
\bottomrule
\toprule
$x$: & Do \textcolor{red}{you} have any reminders for me? \hfill \textbf{(b)} \\
$x'_1$: & do \textcolor{red}{I} have any reminders for today? \\
$y'_1$: & [IN get reminder = \textcolor{red}{[SL person reminded = I]} \\
& \quad [SL date time = for today]] \\
$x'_2$: & Do \textcolor{red}{I} have any reminders for today? \\
$y'_2$: & [IN get reminder = \textcolor{red}{[SL person reminded = I]} \\
& \quad [SL date time = for today]] \\ \midrule
T5: & [IN get reminder = [SL person reminded = me]]  \checkmark \\
\casperOrigAbbr: & [IN get reminder = \textcolor{red}{[SL person reminded = you]}] \\
\bottomrule
\toprule
$x$: & any \textcolor{red}{news updates}? \hfill \textbf{(c)}\\
$x'_2$: & any updates on the news \\
$y'_2$: & [IN get stories news = [SL news type = updates] \\
& \quad [SL news type = news]] \\
$x'_4$: & Are there any \textcolor{red}{news updates} \\
$y'_4$: & [IN get stories news = [SL news type = \textcolor{red}{news updates}]] \\
\midrule
T5: & [IN get stories news = [SL news type = \textcolor{red}{news}]]  \checkmark \\
\casperOrigAbbr: & [IN get stories news = [SL news type = \textcolor{red}{news updates}]] \\
\bottomrule
\end{tabular}
}
\caption{Example predictions by T5 and \casperOrig (\casperOrigAbbr). (2 out of 5 exemplars are shown; \checkmark = correct)}
\label{fig:example-predictions}
\end{figure}

We start with evaluating \casper on the standard train-test setup of the English portion of the MTOP dataset \cite{li-etal-2021-mtop}.\footnote{Available at \url{https://fb.me/mtop_dataset}}
We show that the retrieved exemplars can aid the seq2seq generator even on in-distribution queries.

\subsection{Setup}

\paragraph{Data}
The MTOP dataset uses the decoupled TOP representation \cite{gupta-etal-2018-semantic-parsing,li-etal-2021-mtop},
as exemplified in Figures~\ref{fig:pull-figure},~\ref{fig:improving-faithfulness},~and~\ref{fig:example-predictions}.
A TOP representation is a tree, with each node labeled with either an intent (e.g., IN:CREATE\_CALL) or a slot (e.g., SL:CONTACT). Each node also corresponds to a single token span of the query. The topmost node is always an intent node.

For our models, we simply treat the TOP representation as a string. We start with the string serialization given in the dataset, and then lowercase and word-split the labels to simplify tokenization (e.g., [IN:GET\_CALL \dots] $\to$ [IN get call = \dots]).\footnote{This does not significantly affect the model's accuracy, but it reduces the number of tokens and thus allows more exemplars to fit into the augmented query.}

The English portion of the dataset contains 15667 training, 2235 development, and 4386 test queries. Each query also belongs to one of 11 domains, which will be important in the domain bootstrapping setup (Section~\ref{sec:domain-bootstrapping}).

We define the \emph{template} of a TOP tree to be its tree structure and node labels, with the query tokens discarded (e.g., the template of {[IN:A [SL:B text]]} is {[IN:A [SL:B]]}). In addition to the main evaluation metric of exact match accuracy, we also report template accuracy.

\paragraph{Methods}
The main \casper model is trained on a mixture of original and anonymized training data. We also consider two variants: \casperOrig trained only on original data, and \casperAnon trained only on anonymized data. Since \casperAnon does not know about actual labels, the test data for \casperAnon is also anonymized. None of the models use oracle training data with guiding tag in this section.

\paragraph{Baselines}
We compare against mBART+MT, the best published result from \citet{li-etal-2021-mtop}.
We also consider fine-tuning T5 on the original training data without exemplar augmentation.

\subsection{Results and analysis}

Table~\ref{tab:standard-results} shows the experimental results on the test set of MTOP averaged over 3 runs. The base T5 model already outperforms previous state-of-the-art by 0.7\%. With retrieval-augmentation, \casper further improves upon T5, leading to a total of 1.24\% gain in exact match accuracy.

The \casperOrig variant, which is trained only on non-anonymized data, achieves an even higher gain of 2.1\%. With in-distribution test data, leaning toward memorization rather than following noisy exemplars is likely the best strategy. However, we will show in later sections that \casper trained on mixed data is more robust to out-of-distribution queries and changes in the retrieval index.

Our error analysis shows that, while augmented exemplars improve performance over the baseline in general, they also cause some losses. Figure~\ref{fig:example-predictions}a shows a winning example where \casperOrig is better at predicting the slot that shows up in the augmented exemplars. Figure~\ref{fig:example-predictions}b shows a loss due to the exemplars not being in analogy with the input query, while Figure~\ref{fig:example-predictions}c shows a case where annotation inconsistency between the exemplars and the gold output causes a loss.

Note that while T5 was pre-trained on text data, \casper effectively learns to generate syntactically valid MRs, with only 0.04\% of test outputs being syntactically invalid. Post-hoc filtering or constrained decoding \cite{yin-neubig-2017-syntactic,krishnamurthy-etal-2017-neural} could be used if one needs an absolute guarantee for syntactic correctness.

\subsection{Ablation}\label{sec:ablation}

\paragraph{Retriever}
We train \casperOrig with different choices of the retriever's embedder: BERT-base (embedding of the [CLS] token) \cite{devlin-etal-2019-bert}, BERT-large, and USE-large. We also consider an oracle retriever that only returns examples with the same template as the correct output.

Table~\ref{tab:intrinsic-retrieval-results} reports intrinsic metrics of the retrievers, which include template recall@5 (whether one of the top 5 retrievals has the same template as the gold MR) and label coverage@5 (whether all labels in the gold MR appear among the top 5 retrievals).
Meanwhile, Table~\ref{tab:ablation-results} reports the end-to-end results on the development set of the trained baseline T5 and \casperOrig models. %

We observe that USE-large, being pre-trained on sentence-level tasks, performs better than BERT on both intrinsic and end-to-end evaluation.
On the other hand, the oracle performs much better than USE-large, showing that an improved retriever (e.g., USE-large fine-tuned on the training data) could potentially improve the \casper model.

\paragraph{Exemplar selection}
We compare different ways to select exemplars when constructing training data $(x^+, y)$. The choices include using a fixed top-$k$ list (less diverse) instead of sampled-$k$, and using different number of exemplars $k$.
Note that we always use the top-$k$ list at test time, with the same $k$ as training time.

Table~\ref{tab:ablation-results} compares the results. We see that using sampled exemplars during training and a higher $k$ give additive improvements to the model.
However, note that larger $k$'s can make the augmented query exceed the model's maximum query length.

\begin{table}[t]
\centering
{\small
\begin{tabular}{@{}lcc@{}}
\toprule
\textbf{Retriever} &
\textbf{Template Recall@5} &
\textbf{Label Coverage@5} \\
\midrule
BERT-base & 71.9 & 84.5 \\
BERT-large & 70.2 & 82.9 \\
USE-large & 80.8 & 90.3 \\
oracle & 97.6 & 97.6 \\
\bottomrule
\end{tabular}
}
\caption{Intrinsic evaluation of the retrievers on the development data of MTOP.}
\label{tab:intrinsic-retrieval-results}
\end{table}
\begin{table}[t]
\centering
{\small
\begin{tabular}{@{}llcc@{}}
\toprule
\textbf{Retriever} &
\textbf{Train $x'_i, y'_i$} & 
\textbf{Exact} & 
\textbf{Template} \\
\midrule
USE-large & sampled-5 & 84.67
& 87.98 \\
\midrule
none (baseline T5) & - & 83.18
& 87.22 \\
BERT-base & sampled-5 & 83.83 
& 87.35 \\
BERT-large & sampled-5 & 84.04 
& 87.46 \\
oracle & sampled-5 & 92.39 
& 97.63 \\
\midrule
USE-large & top-1 & 84.06 
& 87.49 \\
USE-large & top-5 & 84.21 
& 87.38 \\
USE-large & sampled-1 & 84.00 
& 87.47 \\
USE-large & sampled-3 & 84.38 & 87.70 \\
USE-large & sampled-10 & 84.79 & 87.96 \\
\bottomrule
\end{tabular}
}
\caption{\textbf{Ablation:} results of \casperOrig variants on the development data of MTOP.}
\label{tab:ablation-results}
\end{table}
\section{Domain bootstrapping experiments}\label{sec:domain-bootstrapping}

In addition to improving the parser on the standard setup,
we will show that by manipulating the retrieval index, we can influence the parser's behavior.
We first consider domain bootstrapping,
where a new domain is being added to a previously trained parser, and we want to quickly update the model using a handful of examples in the new domain.

\subsection{Setup}
In each experiment, we select one domain in MTOP as the new domain to be bootstrapped. Let \Xtrain{} and \Dtrain{} and be the sets of training examples in the \textbf{original} domains and the \textbf{new} domain, respectively. Define  \Xdev{} and \Ddev{} similarly on the development set.

We consider two settings.
In the \emph{seen-bootstrap} setting, at training time,
the parser is given access to \Xtrain{} and a small subset $\Dsupport{} \subseteq \Dtrain{}$ of examples from the new domain.
The parser can choose to fine-tune on \Dsupport{} or use it in anyway it likes.
The parser is then evaluated on \Ddev{} to see how well it produces MRs in the new domain,
as well as on \Xdev{} to ensure that the performance on other domains are not affected.

A more difficult setting is the \emph{unseen-bootstrap} setting: the subset \Dsupport{} is only available to the model at test time and not during training (i.e., \Xtrain{} is the only training data available). A method that cannot incorporate side information from \Dsupport{} at test time would perform poorly in this setting.

\paragraph{Methods}

In both settings, we put all examples available at the time into the retrieval index. For the seen-bootstrap setup, the index contains $\Xtrain{} \cup \Dsupport{}$ at all time. Training examples are constructed from $\Xtrain{}$ and $\Dsupport{}$ with 50\% chance of picking each set.
For the unseen-bootstrap setup, the index contains just $\Xtrain{}$ during training, and $\Dsupport{}$ is added on top at test time. Training examples are constructed from $\Xtrain{}$.

Note that when evaluating on \Ddev{}, exemplars can come from both \Xtrain{} and \Dsupport{} in the retrieval index. If the retriever does its job well, we would still mostly get exemplars from \Dsupport{}.

\paragraph{Baselines}
While there are previous works on domain bootstrapping for semantic parsing without fine-tuning \cite{hou-etal-2020-shot,Zhu2020VectorPN,krone-etal-2020-learning,henderson-vulic-2021-convex},
most of them rely on token-level matching and sequence tagging, which are not directly applicable to the hierarchical MRs from MTOP.
We thus compare \casper with T5, which represents generic seq2seq parsers.

For the unseen-bootstrap setting,
we additionally try fine-tuning T5 on either $\Dsupport{}$ or $\Xtrain{} + \Dsupport{}$ 
for a small number of steps. These \emph{fast update} experiments demonstrate the trade-off of spending additional resources for fine-tuning at test time.

\subsection{Results}

In Table~\ref{tab:domain-bootstrapping-results}, we report results of the models trained in the standard setup (full training data) and the two domain bootstrapping setups (with \Dsupport{} = 100 random examples from \Dtrain{}).
The results are averaged over 5 bootstrapped domains: alarm, calling, event, messaging, and music.

We observe that \casper shows larger improvements upon T5 in the domain bootstrapping settings than the standard setting, ranging from +2\% when $\Dsupport{}$ is seen during training (seen-bootstrap), to +38\% when $\Dsupport{}$ is only available at test time (unseen-bootstrap).
The results show that by modifying the retrieval index, we can change the behavior of \casper without needing to fine-tune on examples from the new domain.

On the unseen-bootstrap setup, the model has to rely solely on the exemplars for unseen semantic labels and parse patterns.
The anonymized training data proves to be crucial for making the model more faithful toward the exemplars, as evidenced by \casper trained on mixed data improving upon \casperOrig, and \casperAnon winning over \casperOrig by a large margin.

\paragraph{Comparison with fast update}
The line plots in Figure~\ref{fig:fast-update} track the accuracy on \Ddev{} and \Xdev{} (bootstrapped domain = alarm) when T5 trained on \Xtrain{} is fine-tuned on the support set for a few iterations at test time.
If only the support set is used for fast update (blue), the model eventually suffers from catastrophic forgetting and degrades on \Xdev{}.
Mixing in \Xtrain{} during fast update (red) solves this issue.
T5 eventually surpasses the unseen-bootstrap \casper (green) on \Ddev{} after processing 512 examples, at which point much more computational resource was already consumed than \casper.

\begin{table}[t]
\centering
{\small
\begin{tabular}{@{}lc@{\;\;}cc@{\;\;}cc@{\;\;}c@{}} \\
\toprule
\textbf{Setting:} &
\multicolumn{2}{@{}c@{}}{\textbf{standard}} &
\multicolumn{2}{@{}c@{}}{\textbf{seen-boot.}} &
\multicolumn{2}{@{}c@{}}{\textbf{unseen-boot.}} \\
\textbf{Train data:} &
\multicolumn{2}{@{}c@{}}{\Xtrain{} + \Dtrain{}} &
\multicolumn{2}{@{}c@{}}{\Xtrain{} + \Dsupport{}} &
\multicolumn{2}{@{}c@{}}{\Xtrain{}} \\
\textbf{Eval data:} &
\Ddev{} & \Xdev{} &
\Ddev{} & \Xdev{} &
\Ddev{} & \Xdev{} \\
\midrule
T5 &
87.63 & 82.83 &
70.70 & 82.73 & \phantom{0}5.65 & 82.73 \\
\casper &
88.61 & 83.62 &
72.74 & 83.73 & 43.90 & 83.87 \\
\casperOrig &
\textbf{89.37} & \textbf{84.24} & 
\textbf{73.32} & \textbf{83.88} & 39.15 & \textbf{84.07} \\
\casperAnon &
\underline{84.47} & \underline{78.98} & 
\underline{63.06} & \underline{79.09} & \textbf{53.79} & \underline{79.22} \\
\bottomrule
\end{tabular}
}
\caption{\textbf{Domain bootstrapping:} exact match accuracy averaged over 5 choices of new domains. On the unseen-bootstrap setting, \casper has the highest gain on the new domain without hurting other domains.  (\underline{underlined} = worse than baseline)}
\label{tab:domain-bootstrapping-results}
\end{table}
\begin{figure}
\centering
\begin{tikzpicture}
\begin{axis}[
    width=7cm,
    height=5cm,
    ymin=74, ymax=85,
    ylabel={accuracy on \Xdev{}},
    axis y discontinuity=crunch,
    ytick distance=2,
    ytickmin=76,
    xmin=0, xmax=80,
    xlabel={accuracy on \Ddev{}},
    legend style={
        font=\tiny,
        legend pos=south west,
        legend cell align=left,
    },
]
\addplot[color=blue,mark=+]
coordinates {
(0.00, 83.16)
(40.32, 82.53)
(56.45, 81.80)
(63.44, 79.11)
(68.82, 78.38)
(69.35, 77.50)
(69.89, 77.31)
(69.89, 76.77)
(67.74, 76.53)
(69.35, 76.23)
};
\addplot[color=red,mark=+]
coordinates {
(0.00, 83.16)
(50.00, 82.58)
(58.60, 82.97)
(61.29, 82.87)
(67.20, 82.77)
(67.20, 82.77)
(68.28, 82.77)
(67.74, 82.53)
(68.28, 82.53)
(69.35, 82.48)
};
\addplot[color=green!70!black,mark=otimes]
coordinates {(63.98, 84.14)};
\legend{
T5: $\Xtrain{}\rightarrow\Dsupport{}$,
T5: $\Xtrain{}\rightarrow\Xtrain{}+\Dsupport{}$,
\casper: $\Xtrain{}$,
}
\end{axis}
\end{tikzpicture}
\caption{
\textbf{Fast update for domain bootstrapping:} accuracy on $\Ddev{}$ and $\Xdev{}$ (new domain = alarm) when T5 trained on \Xtrain{} is fine-tuned on either \Dsupport{} ({\color{blue}blue}) or $\Xtrain{} + \Dsupport{}$ ({\color{red}red}) at test time.
}
\label{fig:fast-update}
\end{figure}

\section{Parse guiding experiments}\label{sec:parse-guiding}

In this section, we demonstrate \casper's ability to guide the prediction toward the patterns specified in the exemplars.
This parse guiding ability can be useful for correcting the parser's output on a set of problematic queries (e.g., sensitive queries, or queries that the model struggles on).
In industrial semantic parsers,
one common way to handle problematic queries is to add explicit ``hotfix'' filters and treat such queries as special cases. Parse guiding enables us to also handle queries that are \emph{sufficiently similar} to known problematic queries. Concretely, we can use the similarity score from \casper's retriever to identify whether the input is similar to any problematic examples, and apply parse guiding toward them if it is.

\subsection{Setup}

We focus on the usage of guiding tag (Section~\ref{sec:faithfulness}) for parse guiding. Trained correctly, the parser should become more faithful toward the exemplars when the guiding tag is present in the augmented query, and should parse normally otherwise.

To evaluate this parse guiding ability, we define an \emph{oracle evaluation set} consisting of examples $(x, E, y)$ with a predefined list of exemplars $E$. The MRs $y'_i$ in $E$ are restricted to having the same semantic template as $y$. On this evaluation, we expect the model's accuracy to rise when the guiding tag is present.
The template accuracy, which is now equivalent to the rate where the prediction follows the template of $y'_i$, should also increase.

\paragraph{Methods}
We compare \casper that was taught the behavior of the guiding tag against the models without such knowledge. We report the results on the standard and oracle evaluation sets, with and without the guiding tag added at test time.

\subsection{Results}

Table~\ref{tab:parse-guiding-results} shows the experimental results.
On the standard evaluation set, \casper model with the knowledge of guiding tag has a slightly smaller gain over T5. But when the guiding tag is present, the model becomes much more faithful to the given exemplars, as evidenced by the increased template and exact match accuracy on the oracle set.

Note that this gain is due to the guiding tag and not just the increased amount of training data: if we add oracle training data \emph{without} guiding tag when training \casper, the accuracy on the oracle set (90.74) is not as high as when we use the guiding tag (93.02). We also note that the guiding tag should only be used when the correct parse is expected to closely follow the exemplars. Using the tag on the standard set hurts the accuracy.

\begin{table}[t]
\centering{\small
\begin{tabular}{@{}l@{\;}cccc@{}}
\toprule
& \textbf{+ tag at} & \textbf{Standard} & \multicolumn{2}{c}{\textbf{Oracle}} \\
\textbf{Method} &
\textbf{test time} &
\textbf{Exact} &
\textbf{Exact} &
\textbf{Template} \\
\midrule
T5 & - & 83.18 & - & - \\
\casper & - & 84.29 & 88.18	& 91.96 \\
+ oracle train
& no
& 83.91 & 89.26	& 93.19 \\
& yes
& \underline{80.58} & \textbf{93.02} & \textbf{97.74} \\
\bottomrule
\end{tabular}}
\caption{\textbf{Parse guiding:} exact match and template accuracy on the standard and oracle evaluation sets.
(\underline{underlined} = worse than baseline)}
\label{tab:parse-guiding-results}
\end{table}
\begin{figure}[t]
\centering
{\small
\begin{tabular}{@{}r@{ }l@{}}
\toprule
$x$: & call Nicholas and Natasha \hfill \textbf{(a)} \\
$x'_2$: & \emph{PLATINUM} How do you make chicken spaghetti \\
$y'_2$: & [IN get recipes = \\
& \quad [SL recipes included ingredient = chicken] \\
& \quad [SL recipes dish = spaghetti]] \\
\midrule
Gold: & [IN create call = [SL contact = Nicholas] \\
& \quad [SL contact = Natasha]] \\
\casperOrigAbbr: & [IN get recipes = \\
& \quad [SL recipes included ingredient = Nicholas] \\
& \quad [SL recipes included ingredient = Natasha]] \\
\bottomrule
\toprule
$x$: & What's the work address with zipcode  \hfill \textbf{(b)} \\
& \quad where James work? \\
$x'_1$: & \emph{PLATINUM} create alarm for 6h35 \\
$y'_1$: & [IN create alarm = [SL date time = for 6h35]] \\
\midrule
Gold: & [IN get location = [SL contact = James]] \\
\casperOrigAbbr: & [IN create alarm = [SL location = \\
& \quad [IN get location = [SL contact = James]]]] \\
\bottomrule
\end{tabular}
}
\caption{Predictions of \casperOrig (\casperOrigAbbr) trained on the mix of standard and oracle training data when given adversarial exemplars. (1 out of 5 exemplars are shown; \emph{PLATINUM} is the guiding tag.)}
\label{fig:adversarial-predictions}
\end{figure}

When the guiding tag is present, the model needs to balance between being faithful to the exemplars and generating a sane parse.
As an analysis,
we try supplying exemplars with a drastically different template from the gold parse.
The first example from Figure~\ref{fig:adversarial-predictions} shows how the model attempts to fit the two names from the query as two slot values.
The second example shows how the model refuses to predict a {SL:DATE\_TIME} slot, despite the guiding tag being present,
since the query does not contain a suitable value for such a slot.
\section{Schema refactoring experiments}\label{sec:schema-refactoring}

In this section, we show how \casper can adapt to changes in the semantic schema.
Although the solution involves modifying the retrieval index like domain bootstrapping (Section~\ref{sec:domain-bootstrapping}),
schema refactoring presents a new challenge: the parser now needs to produce a different output for \emph{in-domain} queries, and must resist the urge to produce semantic labels it has learned during training.

\subsection{Setup}

We consider a \emph{label splitting} scenario where some semantic labels split into two labels each at test time. Following \citet{Gaddy2020OvercomingCD}, we simulate the scenario backward by using the original dataset as post-refactoring data, and merge 10 pairs of similar labels (listed in Appendix~\ref{app:renamed}) to form the pre-refactoring data.
About 35\% of development examples contain at least one label involved in label splitting, and about half of which have their MRs altered after refactoring.

\paragraph{Methods}
At test time, we replace the retrieval index with post-refactoring training data. For models with the knowledge of guiding tag, we add the guiding tag whenever a retrieved exemplar contains a label involved in label splitting.

\subsection{Results}

Table~\ref{tab:schema-change-results} shows the exact match accuracy on the original and refactored development sets. The baseline T5, which cannot incorporate the changed schema, suffers a 13.7\% drop in exact match accuracy.
The \casperOrig model, which leans toward memorizing labels more than utilizing exemplars, has a modest improvement upon T5.
Mixing in anonymized examples during training and using the guiding tag make \casper achieve a high post-refactoring accuracy, while also maintaining the pre-refactoring accuracy compared to the baseline T5.

\begin{table}[t]
\centering
{\small
\begin{tabular}{@{}lcc@{}}
\toprule
\textbf{Method} &
\textbf{Pre-refactoring} &
\textbf{Post-refactoring} \\
\midrule
T5 & 83.27 & 69.59 \\
\casper & 84.52 & 81.21  \\
\casperOrig & 83.50 & 78.52 \\
\midrule 
\multicolumn{3}{@{}c@{}}{\textbf{Models with knowledge of guiding tag}} \\
\casper & 83.89 & \textbf{81.56} \\
\casperOrig & 84.34 & 79.72  \\
\bottomrule
\end{tabular}
}
\caption{\textbf{Schema refactoring:}
Both mixing in anonymized training data and using guiding tags help \casper achieves the best post-refactor accuracy without hurting the pre-refactor accuracy.
}
\label{tab:schema-change-results}
\end{table}
\section{Related works}\label{sec:related-works}

\subsection{Methods for few-shot tasks}

\casper belongs to the family of methods that adapt to new labels, domains, or tasks based on a handful of examples.
If such examples are available at training time,
one could fine-tune the model on them, in which case data augmentation \cite{jia-liang-2016-data,kumar-etal-2019-closer,andreas-2020-good,Lee2021Ex2}
can be used to amplify the data for the new task.
When the few-shot examples are only available at test time, the task is more difficult, and common approaches in the literature include
\emph{metric learning}, \emph{fast update}, and \emph{exemplar augmentation}.

\paragraph{Metric learning}
The main idea of metric learning
\cite{Koch2015SiameseNN,Vinyals2016MatchingNF,Snell2017PrototypicalNF}
is to learn a representation of objects (either inputs or labels) such that objects in the same class are closer together.
Test inputs are then matched to the representation of the few-shot labels or their exemplars.

Metric learning was first applied on  classification tasks \cite{Fritzler2019FewshotCI,sun-etal-2019-hierarchical,zhang-etal-2020-mzet}. Subsequent studies extended metric learning to sequence labeling and semantic parsing by either matching tokens \cite{hou-etal-2020-shot,Hou2020FewJointAF,Zhu2020VectorPN,krone-etal-2020-learning} or matching spans \cite{henderson-vulic-2021-convex,yu-etal-2021-shot}.
However, such rigid notions of substructure matching do not lend themselves to complex hierarchical outputs.
In \casper, the retriever performs \emph{query-level} matching to retrieve exemplars. While the exemplars may not be exactly in the same class as the query, the generator can implicitly reason with them when making predictions. This allows us to generate complex outputs while still gaining benefits from metric learning.

\paragraph{Fast update}
Given the few-shot examples, one could spend a small amount of resource to fine-tune on them for a few training steps.
This creates a trade-off between the amount of resource spent and the performance on the new task.
A common way to improve this trade-off is via \emph{meta learning} \cite{Finn2017MAML,Ravi2017MetaLearnerLSTM,Li2017MetaSGD}. The main idea is to simulate fast update scenarios during training, and update the model's parameters so that the model performs fast updates more efficiently.
Fast update with meta learning has been applied to NLP models for generalizing to unseen tasks or domains \cite{gu-etal-2018-meta,dou-etal-2019-investigating,bansal-etal-2020-learning,chen-etal-2020-low,athiwaratkun-etal-2020-augmented,wang-etal-2021-meta}.

Since fast update explicitly minimizes the loss on the few-shot examples, the updated model is more likely to be faithful toward them, whereas \casper requires additional techniques to increase faithfulness toward exemplars (Section~\ref{sec:faithfulness}).
Nevertheless, \casper has several advantages over fast update.
For instance, while fast update needs to save the information about new labels into the parameters and recall it when parsing test queries, \casper can directly access the new labels in the exemplars when parsing test queries.
Compared to meta learning, training \casper is also much simpler, only requiring off-the-shelf seq2seq fine-tuning.
Finally, while fast update requires the new data to be input-output pairs to fine-tune on, \casper's exemplars can technically be any information (e.g., new semantic schema) that can be augmented to the query.


\paragraph{Exemplar augmentation}
Our work is not the first to use exemplar augmentation for few-shot tasks  \cite{Radford2019LanguageMA,Zhao2021CalibrateBU}.
The most prominent previous work is GPT-3 \cite{Brown2020GPT3}, which can perform new tasks by augmenting exemplars or task description to the query, even without further fine-tuning the model to specifically handle such augmented queries.


Recent works also consider retrieving exemplars from a retrieval index:
\citet{Liu2021WhatMG} retrieves examples for prompting GPT-3,
while \citet{Gupta2021RETRONLU} analyzes how the choices of exemplars and augmentation methods affect a generative semantic parser.
While these works focus on improving generative models on the standard evaluations,
our work proposes how to use retrieval augmentation for controlling the behavior of the generator, which leads to novel use cases (domain bootstrapping, parse guiding, schema refactoring) on top of achieving state-of-the-art on the standard evaluation.

\subsection{Discussion}

\paragraph{Issues in domain bootstrapping} The most straightforward method to adapt a neural model to new domains is to fine-tune it on new training examples. However, this approach not only has a high computation cost, but also suffers from two critical issues.
One is \emph{catastrophic forgetting}: the inability to preserve previous knowledge  \cite{mccloskey1989catastrophic,goodfellow2013empirical}. The other is \emph{model churn}: instability of model predictions on individual examples after fine-tuning. 
Existing work commonly tackles catastrophic forgetting via incremental training, such as imposing constraints on the distance between new and old models \cite{sarwar2019incremental, rosenfeld2018incremental} or jointly learn a generator to reply past examples for training \cite{hu2018overcoming}.
Another existing approach is to identify conflicting data to improve robustness of model updates \cite{Gaddy2020OvercomingCD}.
In \casper, having the retrieval index that stores training examples mitigates catastrophic forgetting by design. And since the model can be controlled without fine-tuning, model churn is reduced. 

\paragraph{Retrieval-augmented generation}
Recent studies have shown the effectiveness of retrieval augmentation in many generative NLP tasks.
How the model actually uses the retrieved information differs among the methods.
Some methods, like \casper,
encode the retrievals alongside the query
and let the model decides how to use them \cite{guu-etal-2018-generating,Hashimoto2018ARF,He2020LearningSP,weston-etal-2018-retrieve,pandey-etal-2018-exemplar,Lewis2020RAG}.
Some utilize alignments between the retrieved examples and the input \cite{sumita-iida-1991-experiments,Gu2018SearchEG,lu-etal-2019-look}.
And some use the retrievals to explicitly manipulate the token scores at each decoding step \cite{zhang-etal-2018-guiding,hayati-etal-2018-retrieval,peng-etal-2019-text,Khandelwal2020GeneralizationTM,Khandelwal2021NearestNM}.



\paragraph{Controllable generation} Several works on controllable generation make use of conditional VAEs, where the latent variable conditioned on the input is the indicator for controlling the output \cite{Hu2017TowardCG,Shen2018ImprovingVE,Zhang2019ImproveDT,Song2019ExploitingPI,shu-etal-2020-controllable}. Other types of control indication include special input tokens \cite{Keskar2019CTRLAC,Dathathri2020PlugAP} or using another neural model as a style discriminator during decoding \cite{Krause2020GeDiGD}.
Our work use \emph{exemplars} as the indicator for controlling the prediction.

\section{Conclusion}

We have presented \casper, a retrieval-augmented semantic parser that uses the retrieved exemplars to influence the predictions. By manipulating the retrieval index and how the exemplars are augmented, we can control the parser's behavior, which is helpful for domain bootstrapping, parse guiding, and schema refactoring.

Future works include fine-tuning the retriever, possibly jointly with the generator, which has potential to improve the model (see Section~\ref{sec:ablation});
introducing more fine-grained control on the faithfulness toward exemplars than the presence/absence of guiding tag;
and pre-training the model on external resources to increase generalization.
\section{Ethical considerations}

This paper proposes a retrieval-augmented semantic parser, the predictive behavior of which can be changed by editing the retrieval index or how the retrieval-augmented query is constructed.
These modifications can only be carried out by the developer of the parser, and not by the users who issue the queries.
The intended use cases of our work include:
(1) adding support for new query domains to the parser;
(2) overriding predictions of a subset of queries, such as sensitive queries or queries that the parser struggles on; and
(3) adapting the parser to an updated semantic schema.

Our method reduces the computational resources needed to retrain the model when enacting the new behavior. That said, the parser needs to be initially trained to recognize retrieval-augmented queries (which can be expensive), and retraining would be required for drastic changes in behavior (e.g., renaming multiple high-frequency semantic labels at once).

While the experiments were done on the English portion of the MTOP dataset, the method is generic to the language of the queries and meaning representations.
Note that the model performance would depend on whether the underlying pre-trained retriever and generator models support the target languages well.

\section*{Acknowledgements}
We want to thank Pete Shaw, Emily Pitler, and the reviewers for helpful comments and suggestions.

\bibliographystyle{acl_natbib}
\bibliography{anthology,reference}

\appendix
\section{Training details}

\paragraph{Data preprocessing}
We used the entire English portion of the  MTOP dataset \cite{li-etal-2021-mtop}, which contains 15667 training, 2235 development, and 4386 test queries. For the input queries, we space-concatenated the tokens from the official tokenization. The output MRs from the dataset file were preprocessed according to the description in Section 3.1: the intent and slot labels are lowercase and word-split at underscores (e.g., [IN:CREATE\_CALL \dots] becomes [IN create call = \dots]). We also removed spaces before the ``]'' tokens.

\paragraph{Retriever}
To embed the queries, we used pre-trained Universal Sentence Encoder \cite{Cer2018UniversalSE}.
Specifically, we used the large version of the encoder.\footnote{\url{https://tfhub.dev/google/universal-sentence-encoder-large/5}}
The embedder was kept fixed.
We computed the embeddings for all queries on CPU. For each query, we cached 100 exemplars with the least cosine embedding distance from the query. The selection of top-$k$ and sampled-$k$ exemplars were only done on these cached exemplars.

In actual deployment, brute force retrieval might be too slow, and fast nearest neighbor methods \cite{Johnson2021BillionScaleSS,Guo2020SCANN} could be used to speed up the retriever.

\paragraph{Training}
For each original example $(x, y)$, we generated 20 lists $E$ of sampled-5 exemplars,
and saved them to dataset files for fine-tuning the T5 \cite{Raffel2020T5} generator model.
We used the base version of the model (220M parameters).
We selected reasonable hyperparameter values and performed some minimal hyperparameter tuning.
Specifically, we used a batch size of 4096 and the learning rate of 0.001.
Training is done for 2000 steps, with early stopping based on the exact match accuracy on the development data. We fine-tuned T5 on 32 Cloud TPU v3 cores. Training takes approximately 2.5 hours.

We ran the experiments on 3 random seeds.
One exception is the domain bootstrapping experiments, where we ran on 1 seed for each of the 5 domains and averaged the results.

\paragraph{Fast update}
For the fast update experiments in Section~\ref{sec:domain-bootstrapping}, we start from the T5 model fine-tuned on \Xtrain{}, and then continue to fine-tune it on either \Dsupport{} or an equal mix of $\Xtrain{}$ and $\Dsupport{}$ (i.e., 50\% chance of picking an example from $\Xtrain{}$; 50\% chance of picking an example from $\Dsupport{}$). We use a batch size of 128 here instead of 4096. Since $|\Dsupport{}| = 100$, each iteration goes over the support set approximately once when fine-tuning on $\Dsupport{}$.

\section{Detailed experimental results}

\begin{table*}[p]
\centering
{\small
\begin{tabular}{@{}lc@{ }lc@{ }lc@{ }lc@{ }l@{}}
\toprule
& \multicolumn{4}{@{}c@{}}{\textbf{Dev}}
& \multicolumn{4}{@{}c@{}}{\textbf{Test}} \\
\textbf{Method} &
\multicolumn{2}{@{}c@{}}{\textbf{Exact}} &
\multicolumn{2}{@{}c@{}}{\textbf{Template}} &
\multicolumn{2}{@{}c@{}}{\textbf{Exact}} &
\multicolumn{2}{@{}c@{}}{\textbf{Template}} \\
\midrule
T5 
& 83.18 & $\pm$0.12 & 87.22 & $\pm$0.07
& 85.06 & $\pm$0.25 & 88.70 & $\pm$0.15 \\
\casper
& 84.29 & $\pm$0.25 & 87.65 & $\pm$0.31
& 85.54	& $\pm$0.09 & 89.13 & $\pm$0.15 \\
\casperOrig
& 84.67 & $\pm$0.13 & 87.98 & $\pm$0.12
& 86.36 & $\pm$0.12 & 89.65 & $\pm$0.08 \\
\casperAnon
& 79.61 & $\pm$0.17 & 82.60 & $\pm$0.00
& 80.85 & $\pm$0.05 & 83.90 & $\pm$0.21 \\
\bottomrule
\end{tabular}
}
\caption{Detailed results on the standard setup (averaged over 3 runs; $\pm$ standard deviation).}
\label{tab:standard-results-detailed}
\end{table*}
\begin{table*}[p]
\centering{\small
\begin{tabular}{@{}l@{\;}cc@{ }lc@{ }lc@{ }l@{}}
\toprule
& \textbf{+ tag at} &
\multicolumn{2}{c}{\textbf{Standard}} & \multicolumn{4}{c}{\textbf{Oracle}} \\
\textbf{Method} &
\textbf{test time} &
\multicolumn{2}{c}{\textbf{Exact}} &
\multicolumn{2}{c}{\textbf{Exact}} &
\multicolumn{2}{c}{\textbf{Template}} \\
\midrule
T5 & - & 83.18 & $\pm$0.12 & - &   & - &  \\
\casper & - & 84.29 & $\pm$0.25 & 88.18 & $\pm$0.40 & 91.96 & $\pm$0.23 \\
+ oracle train
& no
& 83.91 & $\pm$0.07 & 89.26& $\pm$0.85 & 93.19& $\pm$0.83 \\
& yes
& 80.58& $\pm$0.16 & 93.02& $\pm$0.21 & 97.74 & $\pm$0.23 \\
\bottomrule
\end{tabular}}
\caption{Detailed results on parse guiding (averaged over 3 runs; $\pm$ standard deviation)}
\label{tab:parse-guiding-detailed}
\end{table*}
\begin{table*}[p]
\centering
{\small
\begin{tabular}{@{}lc@{ }lc@{ }l@{}}
\toprule
\textbf{Method} &
\multicolumn{2}{c}{\textbf{Pre-refactoring}} &
\multicolumn{2}{c}{\textbf{Post-refactoring}} \\
\midrule
T5 & 83.27 & $\pm$0.21 & 69.59 & $\pm$0.11 \\
\casper & 84.52 & $\pm$0.09 & 81.21 & $\pm$0.21 \\
\casperOrig & 83.50 & $\pm$0.27 & 78.52 & $\pm$0.16 \\
\midrule 
\multicolumn{5}{@{}c@{}}{\textbf{Models with knowledge of guiding tag}} \\
\casper & 83.89 & $\pm$0.12 & 81.56 & $\pm$0.20 \\
\casperOrig & 84.34 & $\pm$0.18 & 79.72 & $\pm$0.18 \\
\bottomrule
\end{tabular}
}
\caption{Detailed results on schema refactoring (averaged over 3 runs; $\pm$ standard deviation).}
\label{tab:schema-change-detailed}
\end{table*}
\begin{table*}[p]
\centering
{\small
\begin{tabular}{@{}lc@{\;\;}cc@{\;\;}cc@{\;\;}c@{}} \\
\toprule
\textbf{Setting:} &
\multicolumn{2}{@{}c@{}}{\textbf{standard}} &
\multicolumn{2}{@{}c@{}}{\textbf{seen-boot.}} &
\multicolumn{2}{@{}c@{}}{\textbf{unseen-boot.}} \\
\textbf{Train data:} &
\multicolumn{2}{@{}c@{}}{\Xtrain{} + \Dtrain{}} &
\multicolumn{2}{@{}c@{}}{\Xtrain{} + \Dsupport{}} &
\multicolumn{2}{@{}c@{}}{\Xtrain{}} \\
\textbf{Eval data:} &
\Ddev{} & \Xdev{} &
\Ddev{} & \Xdev{} &
\Ddev{} & \Xdev{} \\
\midrule
\textbf{alarm} \\
T5 &
83.87 & 83.21 & 
75.81 & 82.77 & \phantom{0}1.61 & 83.21 \\
\casper &
86.56 & 83.94 & 
77.96 & 84.14 & 63.98 & 84.14 \\
\casperOrig &
87.63 & 84.48 & 
77.96 & 84.53 & 62.37 & 84.38 \\
\casperAnon &
80.11 & 79.60 & 
72.58 & 79.65 & 65.59 & 80.14 \\
\midrule
\textbf{calling} \\
T5 &
94.22 & 81.37 & 
72.04 & 81.37 & 24.92 & 81.53 \\
\casper &
94.53 & 82.27 & 
74.16 & 82.95 & 48.33 & 82.63 \\
\casperOrig &
95.14 & 82.95 & 
74.47 & 82.16 & 43.47 & 82.95 \\
\casperAnon &
90.88 & 77.54 & 
64.13 & 77.65 & 55.93 & 77.81 \\
\midrule
\textbf{event} \\
T5 &
92.68 & 82.77 & 
82.93 & 82.72 & \phantom{0}0.00 & 82.48 \\
\casper &
92.68 & 83.57 & 
82.11 & 83.43 & 68.29 & 83.76 \\
\casperOrig &
92.68 & 84.28 & 
83.74 & 83.71 & 65.04 & 84.23 \\
\casperAnon &
89.43 & 78.84 & 
71.54 & 78.74 & 71.54 & 79.12 \\
\midrule
\textbf{messaging} \\
T5 &
94.89 & 82.27 & 
74.43 & 82.13 & \phantom{0}1.70 & 81.88 \\
\casper &
94.89 & 83.15 & 
77.27 & 83.15 & 30.68 & 83.63 \\
\casperOrig &
96.02 & 83.78 & 
77.27 & 83.29 & 21.02 & 83.58 \\
\casperAnon &
90.91 & 78.48 & 
63.07 & 78.44 & 42.05 & 78.34 \\
\midrule
\textbf{music} \\
T5 &
72.46 & 84.52 & 
48.31 & 84.66 & \phantom{0}0.00 & 84.57 \\
\casper &
74.40 & 85.16 & 
52.17 & 85.01 & \phantom{0}8.21 & 85.21 \\
\casperOrig &
75.36 & 85.70 & 
53.14 & 85.70 & \phantom{0}3.86 & 85.21 \\
\casperAnon &
71.01 & 80.42 & 
43.96 & 80.97 & 33.82 & 80.72 \\
\bottomrule
\end{tabular}
}
\caption{Domain bootstrapping results on each domain.}
\label{tab:by-domain}
\end{table*}

Table~\ref{tab:standard-results-detailed},~\ref{tab:parse-guiding-detailed},~and~\ref{tab:schema-change-detailed} show detailed results of the standard, parse guiding, and schema refactoring experiments.
Table~\ref{tab:by-domain} shows per-domain results for the domain bootstrapping experiments. We note that T5 got a non-trivial accuracy on the no-fine-tuning setting of the \emph{calling} domain. This is because many training queries in the \emph{reminder} domain have IN:CREATE\_CALL, a main intent of \emph{calling}, nested inside (e.g., ``Delete reminder to \emph{call husband}'').

\section{Labels in the schema refactoring experiments}\label{app:renamed}

\begin{table*}[tp]
\centering
{\small
\begin{tabular}{@{}llr@{}}
\toprule 
\textbf{Pre-refactoring Label}
& \textbf{Post-refactoring Label}
& \textbf{Count} \\
\midrule
\texttt{IN:GET\_EVENT}
& \texttt{IN:GET\_EVENT} & 724 \\
& \texttt{IN:GET\_REMINDER} & 335 \\
\midrule
\texttt{SL:TYPE\_RELATION}
& \texttt{SL:TYPE\_RELATION} & 1294 \\
& \texttt{SL:TYPE\_CONTENT} & 330 \\
\midrule
\texttt{IN:GET\_MESSAGE}
& \texttt{IN:GET\_MESSAGE} & 220 \\
& \texttt{IN:GET\_TODO} & 281 \\
\midrule
\texttt{SL:MUSIC\_PLAYLIST\_TITLE}
& \texttt{SL:MUSIC\_PLAYLIST\_TITLE} & 96 \\
& \texttt{SL:MUSIC\_PROVIDER\_NAME} & 265 \\
\midrule
\texttt{SL:RECIPES\_SOURCE}
& \texttt{SL:RECIPES\_SOURCE} & 9 \\
& \texttt{SL:RECIPES\_COOKING\_METHOD} & 234 \\
\midrule
\texttt{IN:SWITCH\_CALL}
& \texttt{IN:SWITCH\_CALL} & 49 \\
& \texttt{IN:UPDATE\_CALL} & 218 \\
\midrule
\texttt{IN:GET\_CONTACT}
& \texttt{IN:GET\_CONTACT} & 1537 \\
& \texttt{IN:GET\_LOCATION} & 217 \\
\midrule
\texttt{IN:SET\_AVAILABLE}
& \texttt{IN:SET\_AVAILABLE} & 52 \\
& \texttt{IN:GET\_AVAILABILITY} & 214 \\
\midrule
\texttt{SL:SCHOOL}
& \texttt{SL:SCHOOL} & 153 \\
& \texttt{SL:EMPLOYER} & 204 \\
\midrule
\texttt{SL:NEWS\_TOPIC}
& \texttt{SL:NEWS\_TOPIC} & 617 \\
& \texttt{SL:NEWS\_CATEGORY} & 201 \\
\bottomrule
\end{tabular}
}
\caption{Labels for the schema refactoring experiments.}
\label{tab:renamed-labels}
\end{table*}

Table~\ref{tab:renamed-labels} lists the affected labels in the schema refactoring experiments (Section~\ref{sec:schema-refactoring}), along with their frequencies in the original training data. Note that the training data contains 15667 examples.

\end{document}